%% file: main.tex
\documentclass[letterpaper]{article} 
\usepackage{aaai2026}
\usepackage{times}  
\usepackage{helvet}  
\usepackage{courier}  
\usepackage[hyphens]{url}  
\usepackage{graphicx} 
\usepackage{natbib}  
\usepackage{caption} 
\usepackage{algorithm}
\usepackage{xspace}
\usepackage{amsmath}
\usepackage{amssymb}
\usepackage{booktabs}
\usepackage{multirow}
\usepackage{algpseudocode}
\usepackage{newfloat}
\usepackage{listings}
\DeclareCaptionStyle{ruled}{labelfont=normalfont,labelsep=colon,strut=off} 
\floatstyle{ruled}
\newfloat{listing}{tb}{lst}{}
\floatname{listing}{Listing}
\title{RegionRAG: Region-level Retrieval-Augmented Generation\\for Visual Document Understanding}
\author{
    Yinglu Li, 
    Zhiying Lu, 
    Zhihang Liu, 
    Yiwei Sun,
    Chuanbin Liu\thanks{Corresponding author.}, 
    Hongtao Xie
}
\affiliations{
    University of Science and Technology of China, Hefei, China \\
    \{lulupig, arieseirack, liuzhihang, syw95\}@mail.ustc.edu.cn, \{liucb92, htxie\}@ustc.edu.cn
}
\usepackage{bibentry}

\newcommand{\modelname}{RegionRAG\xspace}

\begin{document}

\maketitle

\begin{abstract}
Multi-modal Retrieval-Augmented Generation (RAG) has become a critical method for empowering LLMs by leveraging candidate visual documents.
However, current methods consider the entire document as the basic retrieval unit, introducing substantial irrelevant visual content in two ways:
1) Relevant documents often contain large regions unrelated to the query, diluting the focus on salient information; 2) Retrieving multiple documents to increase recall further introduces redundant and irrelevant documents.
These redundant contexts distract the model's attention and further degrade the performance.
To address this challenge, we propose \modelname, a novel framework that shifts the retrieval paradigm from the document level to the region level. During training, we design a hybrid supervision strategy from both labeled data and unlabeled data to pinpoint relevant patches. During inference, we propose a dynamic pipeline that intelligently groups salient patches into complete semantic regions. By delegating the task of identifying relevant regions to the retriever, \modelname enables the generator to focus solely on concise, query-relevant visual content, improving both efficiency and accuracy. Experiments on six benchmarks demonstrate that RegionRAG achieves state-of-the-art performance. It improves retrieval accuracy by 10.02\% in R@1 on average, and boosts question answering accuracy by 3.56\% while using only 71.42\% visual tokens compared with prior methods. 
\end{abstract}
\begin{links}
    \link{Code}{https://github.com/Aeryn666/RegionRAG}
    \link{Extended version}{https://arxiv.org/pdf/2510.27261}
\end{links}

\section{Introduction}
Retrieval-Augmented Generation (RAG) is a powerful paradigm that equips Large Language Models (LLMs) with external knowledge by retrieving relevant context from a dynamic database~\cite{chen2022re, yasunaga2022retrieval, liu2025capability}. 
As RAG achieves significant results on text databases, researchers have shifted their focus more to visual document databases with  complex visual layouts that hinder knowledge retrieval and grounding~\cite{yu2024visrag, faysse2024colpali, dong2025scan, tanaka2025vdocrag}.

\begin{figure}[t]
\centering
\includegraphics[width=0.96\columnwidth]{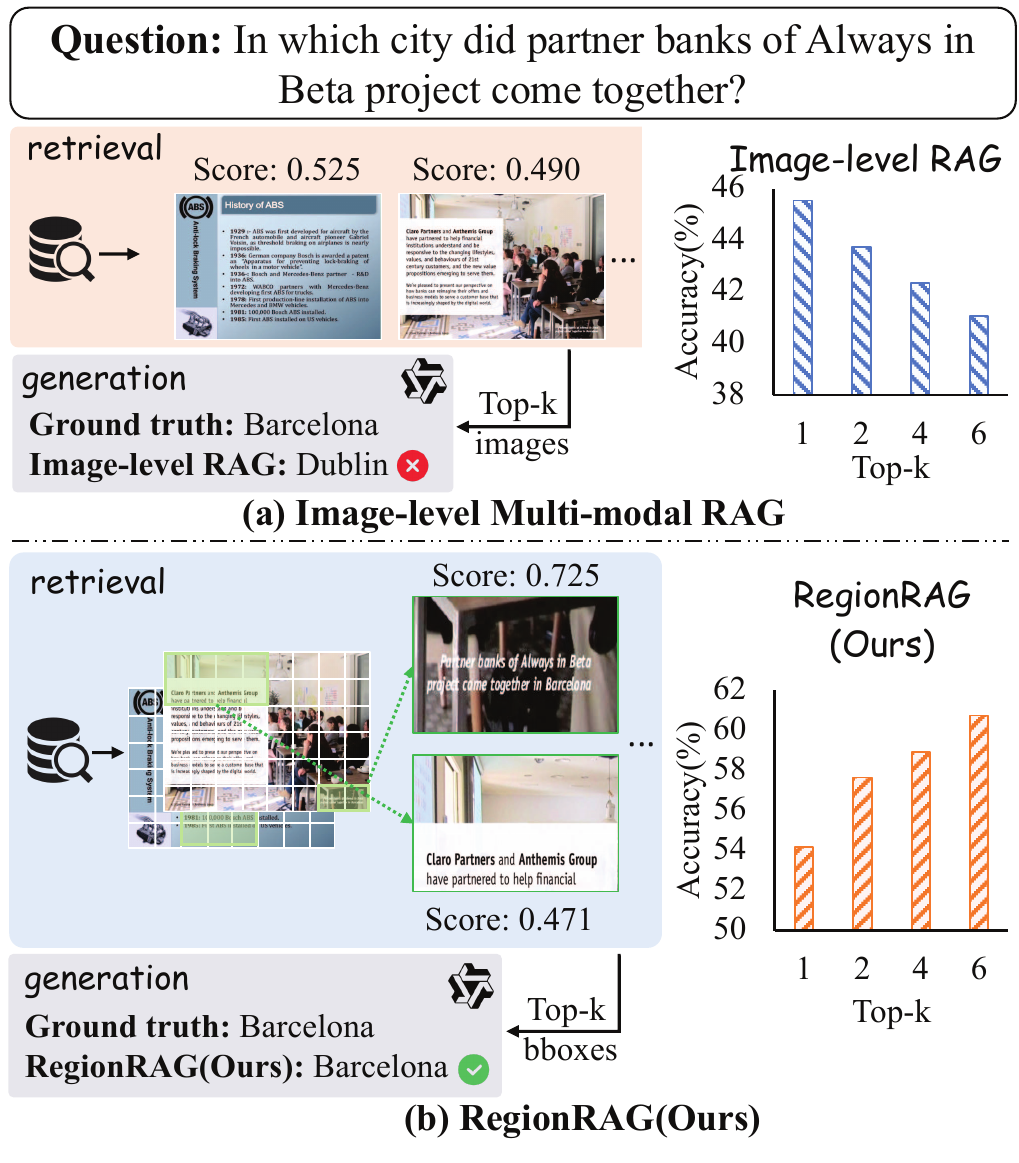}
\caption{Comparison of (a) Image-level RAG with (b) our proposed RegionRAG. While traditional methods retrieve entire yet coarse-grained document images, our RegionRAG identifies and forwards only the most salient regions to the generator. This focused, region-level approach significantly improves final generation accuracy.}
\label{figure1}
\end{figure}

Early visual document RAG approaches relied on brittle pipelines that first extracted text via Optical Character Recognition (OCR) and layout analysis~\cite{zhang2024map}. This process is not only complex but also discards crucial visual information (e.g., layout, style) for holistic comprehension. To address these limitations, the development of Vision-Language Models (VLMs) has spurred a new wave of frameworks that operate directly on document images. Representative works such as VisRAG~\cite{yu2024visrag}, ColPali~\cite{faysse2024colpali}, and VDocRAG~\cite{tanaka2025vdocrag} leverage VLMs to retrieve documents based on holistic visual and textual features, thereby bypassing the error-prone text extraction stage, preserving richer semantics.

Despite recent progress, existing VLM-based RAG systems typically operate at the document level, treating the entire image as a retrieval unit. This coarse granularity introduces two major types of noise that degrade generation performance. The first type is from the irrelevant information in a document. Even relevant documents often contain large regions unrelated to the query, diluting the model's focus on truly salient content. The second type is from the multiple documents specially introduced in RAG. A common strategy is to feed the top-$k$ retrieved documents into the generator~\cite{yao2025spotlight} to compensate for retrieval errors. However, this introduces a performance paradox: while intended to increase ground-truth recall, our experiment shown in Figure~\ref{figure1}(a) reveals that increasing~$k$ actually leads to a decrease in performance. This may be because including redundant and irrelevant documents significantly increases visual tokens and distracts the model's attention, thereby outweighing the limited performance gains provided by the small amount of valid information.

To tackle this problem, an intuition is to shift the retrieval granularity from the entire document to only the most relevant semantic regions. Therefore, we propose a novel framework named RegionRAG that enhances retrieval at both the training and inference stages.
During training, we improve the alignment between queries and document regions through a hybrid supervision strategy. Specifically, RegionRAG jointly leverages manually annotated bounding boxes from labeled data and weakly-supervised similarity signals from unlabeled data. A unified loss integrates both supervision types to learn precise query-patch alignment while minimizing annotation cost. 
During inference, we introduce a dynamic region proposal algorithm based on patch-specific similarity maps. Since relevant information often spans multiple neighboring patches (e.g., parts of a table or paragraph), we apply a neighbor-based grouping strategy that merges patches into coherent regions. This enhances matching precision by isolating concise, context-complete visual segments relevant to the query. 
As shown in Figure~\ref{figure1}(b), our region-level RAG can effectively improve performance as $k$ increases.
Extensive experiments demonstrate that RegionRAG achieves a 3.56\% average accuracy gain across five document visual question answering (VQA) benchmarks, and cuts 28.58\% costs of visual tokens compared to prior methods.

Our main contributions are summarized as follows:
\begin{itemize}
    \item To the best of our knowledge, we are the first to propose a region-level multi-modal RAG framework (RegionRAG) for visual documents.
    \item We introduce a dual-objective training strategy that achieves fine-grained patch alignment using different types of data, and introduce a neighbor-based grouping inference strategy to get visual regions.
    \item Experiments show our RegionRAG achieves more accurate retrieval and efficient generation compared with previous methods.
\end{itemize}

\section{Related Works}
\subsubsection{Image-Level Retrieval-Augmented Generation (RAG).}
RAG enhances Large Language Models (LLMs) by incorporating external knowledge, a paradigm that was initially successful in natural language processing (NLP) tasks~\cite{guu2020retrieval, borgeaud2022improving, ram2023context}. It typically involves retrieving relevant information from an external knowledge base to guide answer generation~\cite{shi2023replug, yu2023augmentation, liu2024towards}. Beyond NLP, RAG has been extended to visually rich documents. Early works often relied on brittle OCR pipelines~\cite{yu2023augmentation}, which discarded visual cues and introduced fragile pre-processing steps. However, this approach discards visual cues and introduces brittle pre-processing steps. Recently, the rapid development of Vision-Language Models (VLMs)~\cite{liu2023visual, liu2025hybrid}, such as VisRAG~\cite{yu2024visrag}, ColPali~\cite{faysse2024colpali}, and VDocRAG~\cite{tanaka2025vdocrag} has enabled the direct retrieval of entire document images. While pioneering, these methods operate at a coarse, document-level granularity. This forces the generator to process entire images, introducing substantial noise from query-irrelevant content, which dilutes contextual relevance, degrades generation quality, and reduces computational efficiency.

\subsubsection{General Visual Document Understanding.}
Another line of research focuses on end-to-end understanding of individual visual documents. Traditional OCR-based pipelines~\cite{zhang2024map} are fragile and discard layout cues crucial for reasoning. With the emergence of multimodal LLMs, models such as DocFormer~\cite{appalaraju2021docformer}, Qwen-VL~\cite{bai2023qwen}, and InternVL~\cite{chen2024internvl} demonstrate unified reasoning over document images. However, these models are not designed for large-scale retrieval. Retrieval-augmented methods like VisRAG~\cite{yu2024visrag} and TabPedia~\cite{zhao2024tabpedia} leverage visual grounding but still retrieve entire documents, introducing redundant context. In contrast, RegionRAG retrieves fine-grained, query-relevant regions for precise and efficient context construction.

\subsubsection{Region-Based VQA and Grounding.}
Inspired by the need for finer granularity, another category of models explores region-level reasoning. Studies such as RegionGPT~\cite{guo2024regiongpt}, region-based VQA methods~\cite{wu2022region}, VLM-R3~\cite{jiang2025vlm}, and DeepEyes~\cite{zheng2025deepeyes} adopt an end-to-end generate-and-ground paradigm. However, their task also differs crucially from ours: they primarily perform localization or grounding within a single, pre-selected document to answer a query. They are not retrieval systems designed to search a corpus. While one could envision an agentic, multi-step pipeline for corpus-level retrieval using these models, such approaches are prohibitively slow for efficient retrieval. In contrast, RegionRAG is the first to formalize an explicit, decoupled region-retrieval stage for the RAG paradigm, bridging the gap between efficient retrieval and fine-grained understanding.

\begin{figure*}[t]
\centering
\includegraphics[width=0.95\textwidth]{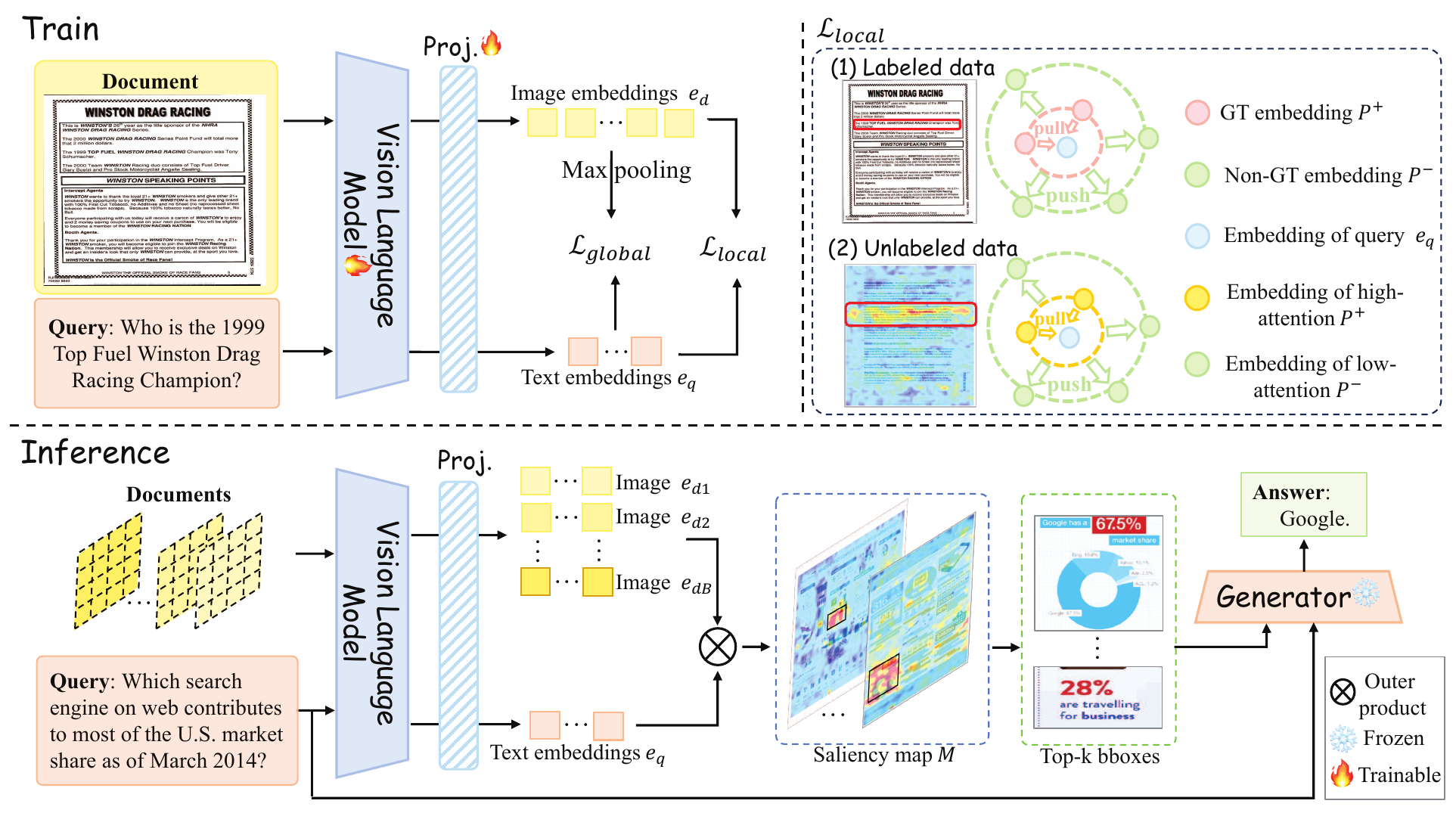} 
\caption{The framework of RegionRAG. During training, the model jointly learns global (document-query) and local (region-query) alignments. At inference time, it first identifies and retrieves the most relevant visual regions and subsequently generates an answer based on this retrieved context.}
\label{figure2}
\end{figure*}

\section{Methods}
\subsection{Preliminary}
Our RegionRAG framework redefines the standard RAG pipeline for visually-rich documents. We begin with a corpus~$\mathcal{C}$ containing~$M$ documents,~$\mathcal{C}=\{D_1,D_2,\dots,D_M\}$. Each document~$D_m$ is segmented into~$N$ non-overlapping patches~$\{p_1,p_2,\dots,p_N\}$. These patches serve as our fundamental retrieval units, contrasting with traditional methods that operate on whole documents.

Our pipeline consists of two main components. Firstly, a RegionRetriever~$(\mathcal{R})$, is responsible for identifying the most relevant set of patches~$P_{\mathcal{R}}$ from a document~$D_m$ given a text query~$q$. It uses a shared VLM to embed both the query and the visual patches into a common latent space for fine-grained similarity matching; its function is represented as~$P_{\mathcal{R}}=\mathcal{R}(q,D_m)$. 

We train~$(\mathcal{R})$ with two levels of alignment to develop its region retrieval ability, a global document-query level and a fine-grained region-query level.
Secondly, we employ a Generator~$(\mathcal{G})$ to synthesize the final textual answer~$a$. As an MLLM, it is conditioned on the original query~$q$ and the set of retrieved image patches~$P_{\mathcal{R}}$; 
this process is denoted as~$a=\mathcal{G}(q,P_{\mathcal{R}})$.

\subsection{Global Document-Query Alignment}
While our ultimate goal is to retrieve specific regions, the model must first identify the correct parent document. Our framework first performs Global Document-Query Alignment to obtain a holistic document representation and ensure basic retrieval ability among documents.

We follow~\cite{faysse2024colpali, yu2024visrag} to employ global contrastive learning. To create a holistic document representation from its constituent patches, we aggregate their features into a single global vector. We begin by segmenting a document image~$D$ into~$N$ patches. A VLM then processes these patches to produce a set of~$N$ embeddings,~$E_D=\{e_1,e_2,\dots,e_N\}$, where each~$e_k\in \mathbb{R}^d$ and~$d$ is the embedding dimension.
We apply an element-wise max-pooling operation across the~$N$ patch embeddings~$E_D$ to obtain the single global representation~$v_D$.
During training, we sample a batch of~$B$ document-query pairs~$\{D_i, q_i\}_{i=1}^{B}$. For each pair, we compute the document embedding~$v_{D_i}$ and the query embedding~$e_{q_i}$. The similarity between the~$i$-th query and the~$j$-th document is measured by cosine similarity:
\begin{equation}
s(e_{q_i}, v_{D_j}) = \frac{e_{q_i} \cdot v_{D_j}}{\left\|e_{q_i}\right\| \left\|v_{D_j}\right\|}.
\end{equation}

The objective function for global contrastive learning is based on InfoNCE loss~\cite{he2020momentum}. For the~$i$-th pair in the batch, we treat~$(e_{q_i},v_{D_i})$ as the positive pair and~$(e_{q_i},v_{D_j})$ for all~$j\neq i$ as the negative pairs. The loss for the entire batch is:
\begin{equation}
\mathcal{L}_{global} = -\frac{1}{b} \sum_{i=1}^{b} \log \left( \frac{\exp(s(e_{q_i}, v_{D_i}) / \tau)}{\sum_{j=1}^{b} \exp(s(e_{q_i}, v_{D_j}) / \tau)} \right),
\end{equation}
where~$\tau$ is a temperature parameter. This global alignment stage leverages inter-document level supervision, enabling the model to retrieve relevant documents from the corpus.

\subsection{Fine-grained Region-Query Alignment}
The coarse-grained global alignment only provides inter-document supervision and cannot achieve precise region-level activation within a document, thus failing to support fine-grained region retrieval. Therefore, we introduce Fine-grained Region-Query Alignment to enable intra-document supervision.
This is primarily implemented through the local objective function~$\mathcal{L}_{local}$, which guides the model to pinpoint the specific regions most relevant to the query. The core of our region-level approach is a hybrid supervision strategy that leverages two distinct signal sources: gold-standard annotations from labeled data (with bounding boxes) and weakly-supervised signals from unlabeled data (without bounding boxes). Both supervision types share a unified loss formulation.

We consider a training batch of~$B$ document-query pairs, denoted as~$\{D_i, q_i\}_{i=1}^{B}$. For each pair, the VLM provides a query embedding~$e_{q_i}\in\mathbb{R}^d$ and a set of~$N$ patch embeddings for the document~$D_i$, denoted as~$E_{D_i} = \{e_{i,1}, e_{i,2}, ..., e_{i,N}\}$. Each patch embedding~$e_{i,k} \in \mathbb{R}^d$ represents a specific document region. The core of our fine-grained stage is to compute the similarity between the query~$e_{q_i}$ and each patch embedding~$e_{i,k}$ to identify the most relevant regions.

\begin{figure}[!t]
\centering
\includegraphics[width=0.85\columnwidth]{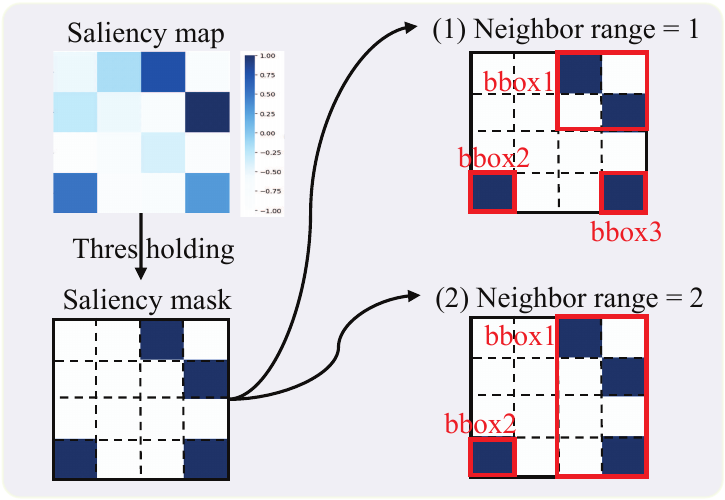}
\caption{Our neighbor-based grouping strategy. Patches with high saliency are binarized and then grouped into bounding boxes.}
\label{figure3}
\end{figure}

\subsubsection{Supervision from Labeled Data.}
For labeled data, each document-query pair~$(D_i, q_i)$ is accompanied by ground-truth bounding boxes that delineate the regions essential for answering the query. We use this precise supervision to construct positive and negative sets of patch embeddings. The positive set,~$P_i^+$, consists of all patch embeddings~$\{e_{i,k}\}$ corresponding to image patches centered within any ground-truth box. Conversely, the negative set,~$P_i^-$, consists of all patch embeddings for patches lying entirely outside these boxes.
The local loss ~$\mathcal{L}_{local}^L$ for a labeled sample encourages the query embedding~$e_{q_i}$ to be closer to the embeddings in~$P_i^+$ than to those in~$P_i^-$. 

\subsubsection{Supervision from Unlabeled Data.}
A key advantage of our method is its ability to learn from unlabeled data, which contains only weakly-supervised document-query pairs~$(D_i, q_i)$ without region annotations. To extract supervisory signals, we generate pseudo-ground-truth labels by leveraging the model's similarity estimates between the query and document patch embeddings. This is feasible because a well-pretrained VLM, guided by the global loss and supervision from labeled data, naturally exhibits some patch-level attention capability.
We compute a saliency map that captures the interaction between the query~$q_i$ and the document patch embeddings~$E_{D_i}$. This map is then binarized using a predefined threshold~$\theta$. To ensure stable selection of positive samples, we adopt a relatively low value for~$\theta$. Patch embeddings with similarity scores above~$\theta$ are included in the positive set~$P_i^+$, while those below are assigned to the negative set~$P_i^-$.

Given the positive set~$P_i^+$ and negative set~$P_i^-$ (derived from either ground-truth or pseudo-labels), the unified contrastive loss~$\mathcal{L}_{\text{local}}$ is formulated as:
\begin{equation}
\mathcal{L}_{\text{local}} = 
-\log \left( 
\frac{
\sum_{p \in P_i^+} \exp(s(e_{q_i}, p) / \tau)
}{
\sum_{p \in \{P_i^+, P_i^-\}} \exp(s(e_{q_i}, p) / \tau)
} 
\right),
\end{equation}
where~$\tau$ is a temperature parameter.

\subsection{Overall objective}
The overall RegionRAG learning objective is a weighted sum of the global and fine-grained alignment loss:
\begin{equation}
    \mathcal{L}_{\text{RegionRAG}}=\alpha\mathcal{L}_{global}+\beta\mathcal{L}_{local},
\end{equation}
where the weights~$\alpha$ and~$\beta$ are hyper-parameters.

\subsection{Inference}
Building upon the model's capacity for local region attention developed during training, our inference process extracts specific areas to perform region-level retrieval and subsequent answer generation. The pipeline unfolds in three steps: first, we obtain a proposal mask from a retrieval saliency map. Second, we identify connected components of salient patches based on a defined neighborhood size. Third, we extract the minimum bounding rectangle for each element. Finally, we feed the retrieved regions into a generator~$\mathcal{G}$ to produce the final answer.

\subsubsection{Region Proposal via Saliency Mapping.}
The inference process begins by identifying candidate regions within a document image~$D$ that are potentially relevant to the input query~$q$. We first compute a saliency map~$S = \{s(e_q, p_k) | k = 1,2,\dots, N \}$, which is a 2D grid of cosine similarity corresponding to the layout of the~$N$ document patches. 
To filter out patches with low relevance, we binarize this saliency map into a mask~$M$ using a predefined threshold~$\eta \in [0,1]$. Patches with scores above this threshold are considered salient candidates.
\begin{equation}
M_k =
\begin{cases}
1, & \text{if } S_k \ge \eta \\
0, & \text{if } S_k < \eta
\end{cases}.
\end{equation}
This step isolates a sparse set of potentially relevant regions, as shown in Figure~\ref{figure3}, preparing them for the subsequent grouping and merging process.

\subsubsection{Region Merging and Bounding Box Generation.}
To prevent excessive fragmentation of semantically related areas, we group salient patches from the mask~$M$ into connected components. Two patches are considered connected if the distance between their spatial coordinates is less than or equal to a neighborhood radius~$r$. We use the Chebyshev distance as our metric:
\begin{equation}
D((x_1, y_1), (x_2, y_2)) = \max(|x_1 - x_2|, |y_1 - y_2|).
\label{Dis}
\end{equation}
The radius~$r$ controls the merging granularity, allowing for either tight or broad region proposals as illustrated in Figure~\ref{figure3}. While exhaustively traversing all patches is inefficient, we employ a Breadth-First Search (BFS) algorithm for efficient traversal. Specifically, we use a queue to manage the search and mark patches as visited to avoid redundant processing. For each connected component found, we compute its minimum bounding rectangle. Therefore, these bounding rectangles can be considered our final retrieved regions, and can be further sent into the generator~$\mathcal{G}$ to complete the question answering. For each image, the entire region retrieval process is summarized in the Algorithm~\ref{alg:bbox_proposal}.

\renewcommand{\algorithmicrequire}{\textbf{Input:}}
\renewcommand{\algorithmicensure}{\textbf{Output:}}
\begin{algorithm}[t]
\caption{Bounding Box Proposal from Saliency Map}
\label{alg:bbox_proposal}
\begin{algorithmic}[1]
\Require Saliency map~$S$, threshold~$\eta$, neighbor range~$r$, image size~$(H_{\text{img}}, W_{\text{img}})$, grid size~$(H_p, W_p)$
\Ensure A set of bounding boxes~$\mathcal{B}$
\State $M \gets S \geq \eta$ \Comment{Binarize saliency map}
\State $\mathcal{R} \gets \text{FindComponentsBFS}(M, r)$ \Comment{Find connected components}
\State $\mathcal{B} \gets \emptyset$
\For{each component $R$ in $\mathcal{R}$}
    \State $B \gets \text{CalculateMinBbox}(R, (H_{\text{img}}, W_{\text{img}}), (H_p, W_p))$
    \State $\mathcal{B} \gets \mathcal{B} \cup \{B\}$
\EndFor
\State \Return $\mathcal{B}$
\end{algorithmic}
\end{algorithm}

\section{Experiments}
\subsection{Experimental Settings}
\subsubsection{Datasets.}
We train our model using a combination of the unlabeled dataset from VisRAG~\cite{yu2024visrag} in-domain data and the document-focused subset of Visual-CoT~\cite{shao2024visual} for labeled, bounding-box level supervision. For evaluation, we test our method on a diverse suite of benchmarks including DocVQA~\cite{tito2023hierarchical}, PlotQA~\cite{methani2020plotqa}, ArxivQA~\cite{li2024multimodal}, InfoVQA~\cite{mathew2022infographicvqa}, SlideVQA~\cite{tanaka2023slidevqa}, and ViDoRe~\cite{faysse2024colpali}. 

\subsubsection{Evaluation Metrics.}
We report the retrieval and generation performance on the evaluation sets of the datasets sourced from the VQA datasets. For retrieval, we use Recall@1, Recall@10, and nDCG@5. For generation, we follow VisRAG~\cite{yu2024visrag} to report the answer accuracy, employing a relaxed extract match metric that allows a 5\% error margin for numeric responses.

\subsubsection{Implementation Details.}
We initialized RegionRetriever with Qwen2.5-VL-3B~\cite{bai2025qwen2} for training, while the Generator is an off-the-shelf model without training. The model is trained for five epochs on a mixture of 98k labeled and 122k unlabeled data. The batch size per GPU is set to 64. The temperature parameter~$\tau$ is set to 0.02 for the global loss and 0.25 for the local loss.

\subsection{Performance Comparison}
We compare our RegionRAG with the following state-of-the-art methods, including OCR-based pipelines BM25~\cite{robertson2009probabilistic}, BGE-large-en-v1.5~\cite{xiao2024c}, and NV-Embed-V2~\cite{lee2024nv}, and VLM-based methods SigLIP~\cite{zhai2023sigmoid}, VisRAG~\cite{yu2024visrag}, VDocRAG~\cite{tanaka2025vdocrag}, ColPali and ColQwen2.5~\cite{faysse2024colpali}.

\subsubsection{Retrieval Performance.}
\begin{table}[!t]
\small
\setlength\tabcolsep{1mm}
\centering
\begin{tabular}{@{}lc|ccccc|c@{}}
\toprule
\textbf{Models} & \textbf{\#Para} & \textbf{Arxiv} & \textbf{Doc} & \textbf{Info} & \textbf{Plot} & \textbf{Slide} & \textbf{Avg}\\ \midrule
BM25 & n.a. & 54.3 & 86.8 & 82.6 & 76.0 & 91.6 & 78.3 \\
SigLIP & 883M & 45.0 & 68.0 & 84.7 & 58.3 & 89.0 & 69.0 \\
BGE(large) & 335M & 48.7 & 68.2 & 88.2 & 73.1 & 90.1 & 73.7 \\
NV-Embed-V2 & 7.85B & 70.1 & 89.9 & 95.1 & 80.9 & 97.8 & 86.8 \\
ColPali$^*$ & 2.92B & 79.4 & 90.2 & 88.9 & 31.4 & 92.8 & 76.5 \\
VisRAG$^*$ & 3B & 87.6 & 91.2 & 97.0 & 89.3 & 97.1 & 92.4 \\
ColQwen2.5$^*$ & 3B & 84.5 & 97.4 & 98.6 & 47.5 & 97.0 & 85.0 \\ \midrule
RegionRAG (ours) & 3B & \textbf{92.5} & \textbf{99.4} & \textbf{99.5} & \textbf{92.4} & \textbf{98.4} & \textbf{96.4} \\ \bottomrule
\end{tabular}
\caption{Retrieval performance comparison between our RegionRAG and other SOTA methods on multiple types of DocumentVQA benchmarks in Recall@10. $*$ denotes that we reproduce the evaluation using their official checkpoints.}
\label{tab:retrieve1}
\end{table}
\begin{table}[!t]
\small
\setlength\tabcolsep{1mm}
\centering
\begin{tabular}{@{}lc|cccccc@{}}
\toprule
\textbf{Models} & \textbf{\#Data} & \textbf{Arxiv} & \textbf{Doc} & \textbf{Info} & \textbf{Plot} & \textbf{Slide} & \textbf{\small ViDoRe}\\ \midrule
\multicolumn{2}{c|}{} & \multicolumn{6}{l}{\small \textit{Recall@1}} \\ 
VisRAG$^*$ & \multicolumn{1}{l|}{362k} & 69.1 & 66.2 & 78.8 & 49.1 & 76.4 & - \\
ColQwen2.5$^*$ & \multicolumn{1}{l|}{128k} & 62.3 & 80.2 & 82.3 & 9.3 & 76.1 & \textbf{83.3} \\
RegionRAG & \multicolumn{1}{l|}{220k} & \textbf{78.4} & \textbf{86.8} & \textbf{88.9} & \textbf{55.3} & \textbf{80.6} & 82.9 \\ \midrule
\multicolumn{2}{c|}{} & \multicolumn{6}{l}{\small \textit{nDCG@5}} \\ 
VisRAG$^*$ & \multicolumn{1}{l|}{362k} & 78.8 & 77.7 & 88.4 & 65.5 & 89.9 & - \\
ColQwen2.5$^*$ & \multicolumn{1}{l|}{128k} & 71.8 & 87.8 & 90.3 & 19.0 & 86.7 & \textbf{89.4} \\
VDocRAG & \multicolumn{1}{l|}{541k}& -&-&72.9&-&77.3&- \\
RegionRAG & \multicolumn{1}{l|}{220k} & \textbf{84.5} & \textbf{93.1} & \textbf{94.8} & \textbf{70.3} & \textbf{90.3} & 88.4 \\ \bottomrule
\end{tabular}
\caption{Retrieval performance comparison in Recall@1 and nDCG@5. $*$ denotes that we reproduce the evaluation based on their official checkpoints.}
\label{tab:retrieve2}
\end{table}
As shown in Table~\ref{tab:retrieve1}, our RegionRAG outperforms existing methods on the Recall@10 metric across all benchmarks. Compared to VisRAG, which has a comparable number of model parameters (3B) and is trained on 362k samples, our model achieves superior performance with less data (220k). On average, RegionRAG surpasses VisRAG by 4.0 percentage points in Recall@10. 
When compared with ColQwen2.5, which also adopts Qwen2.5-VL-3B as its backbone, the advantage of RegionRAG becomes even more pronounced. Its average Recall@10 score is 11.4 percentage points higher (96.4 vs. 85.0), with a huge margin on the Plot benchmark (92.4 vs. 47.5).
These results demonstrate that our proposed method of supervising the most query-relevant image regions is even superior on the whole document retrieval than employing image-level supervision. This success can be attributed to our tailored training scheme and loss design. We further compare RegionRAG with recent VLM-based methods on Recall@1 and nDCG@5 metrics in Table~\ref{tab:retrieve2}. Our model attains leading performance on most benchmarks, highlighting its consistent superiority. While its scores on ViDoRe are marginally behind those of the top performer, this evaluation constitutes a partial zero-shot setting for our model due to the limited subset domain coverage in our training data.

\subsubsection{Region-Text Alignment Capability.}
We evaluate the model's ability to localize query-relevant regions by comparing the average retrieval similarity between the full image and the ground-truth bounding box (Bbox) from the Visual-CoT test split. As shown in Table~\ref{tab:bbox_image}, both ColQwen2.5 and our RegionRAG achieve higher similarity on Bbox regions, indicating that relevant information is locally concentrated.
Notably, RegionRAG yields a larger similarity gain (0.136 vs. 0.039 for ColQwen2.5), highlighting the effectiveness of our region-level contrastive loss in capturing fine-grained, query-relevant visual cues.
\begin{table}[!t]
\footnotesize
\setlength\tabcolsep{1mm}
\centering
{%
\begin{tabular}{@{}lc|cc|c@{}}
\toprule
\multirow{1}{*}{\textbf{Methods}} & \multirow{1}{*}{\textbf{Region}} & \multicolumn{1}{c}{\textbf{DocVQA}} & \multicolumn{1}{c|}{\textbf{InfoVQA}} & \multicolumn{1}{c}{\textbf{Average}}  \\ 
\midrule
\multirow{2}{*}{\centering ColQwen2.5} & Image & 0.062 & 0.109 & 0.086\\
& Bbox  & 0.112 &0.138 & 0.125 \\
\midrule
\multirow{2}{*}{\centering \modelname (Ours)} & Image & 0.119& 0.226 & 0.173\\
& Bbox  & \textbf{0.271} &\textbf{0.346} & \textbf{0.309} \\
\bottomrule
\end{tabular}%
}
\caption{Comparison of average similarity scores between text queries and different visual scopes (full image vs. ground-truth bounding box) on DocVQA and InfoVQA.}
\label{tab:bbox_image}
\end{table}

\subsubsection{Generation Performance.}
\begin{table}[!t]
\setlength\tabcolsep{1mm}
\centering
\begin{tabular}{@{}clccccc@{}}
\toprule
\textbf{\small Pixel} & \multicolumn{1}{c}{\textbf{\small Methods}} & \textbf{\small Arxiv} & \textbf{\small Doc} & \textbf{\small Info} & \textbf{\small Plot} & \textbf{\small Slide} \\ \midrule
\multirow{5}{*}{\rotatebox{90}{256$^\text{2}$}} & top1 image & \textbf{62.25} & 19.97 & 24.23 & 16.69 & 38.49 \\
 & top1 bbox & 62.13 & \textbf{31.13} & \textbf{39.13} & \textbf{17.15} & \textbf{39.92} \\ \cmidrule(l){2-7} 
 & top4 image & 61.40 & 18.95 & 22.42 & 17.73 & 38.31 \\
 & top4 bbox & \textbf{62.62} & \textbf{42.3} & \textbf{44.15} & \textbf{19.00} & \textbf{44.42} \\ \cmidrule(l){2-7} 
 & oracle & 64.71 & 19.97 & 24.37 & 20.39 & 39.39 \\ \midrule
\multirow{5}{*}{\rotatebox{90}{512$^\text{2}$}} & top1 image & 64.46 & 69.20 & 50.97 & \textbf{24.10} & 59.17 \\
 & top1 bbox & \textbf{64.83} &\textbf{70.39} & \textbf{59.19} & \textbf{24.10} & \textbf{62.94} \\ \cmidrule(l){2-7}
 & top4 image & 63.73 & 70.22 & 47.91 & 23.52 & 61.69 \\
 & top4 bbox & \textbf{65.07} & \textbf{74.28} & \textbf{63.79} & \textbf{24.56} & \textbf{66.01} \\ \cmidrule(l){2-7}
 & oracle & 67.77 & 75.63 & 51.39 & 37.78 & 63.31 \\ \midrule
\multirow{5}{*}{\rotatebox{90}{dynamic (top4)}} & image acc & \textbf{65.2} & 70.22 & 46.94 & \textbf{24.10} & 62.77 \\
 & \#tokens & 1258 & 1268 & 1248 & 1266 & 1259 \\ \cmidrule(l){2-7} 
 & bbox acc & 64.83 & \textbf{71.24} & \textbf{63.23} & 23.64 & \textbf{64.03} \\
 & \#tokens & 1035 & 903 & 661 & 966 & 936 \\ \cmidrule(l){2-7} 
 & oracle & 68.01 & 75.63 & 51.39 & 37.78 & 63.31 \\ \bottomrule
\end{tabular}%
\caption{Performance comparison under fixed and dynamic resolutions across five VQA benchmarks. ``topk image/bbox" denotes processing the retrieved full image or bounding box, ``oracle" uses ground-truth images, ``dynamic" means using the original resolution with up to $512^2$.}
\label{tab:generate}
\end{table}
We evaluate our method under both fixed and dynamic resolution settings. In the fixed setup, we process the retrieved region at either $256^2$ or $512^2$ resolution. As shown in Table~\ref{tab:generate}, accuracy improves notably at both resolutions, and using retrieved bboxes consistently outperforms full-image inputs under the same resolution. In the dynamic setting, which better reflects real-world inference, the model uses the original resolution with up to $512^2$. 
On average, the bbox input consumes only 71.4\% of the image tokens while maintaining or even surpassing full-image performance. For instance, it achieves 16.29\% higher accuracy on InfoVQA. Notably, the full-image method is upper-bounded by the ``oracle" setting, which inputs the ground-truth image at a fixed resolution. Interestingly, our bounding box approach can even surpass this baseline. For example, on InfoVQA at $512^2$ resolution, our top-4 bbox method attains 63.79\% accuracy, exceeding the oracle's 51.39\%. This is because downscaling affects smaller cropped regions less severely than entire images—preserving finer details that enable the model to better perceive the relevant content. Overall, these results highlight the advantages of our region retrieval in both efficiency and fidelity.
\begin{figure*}[!t]
\centering
\includegraphics[width=0.95\textwidth]{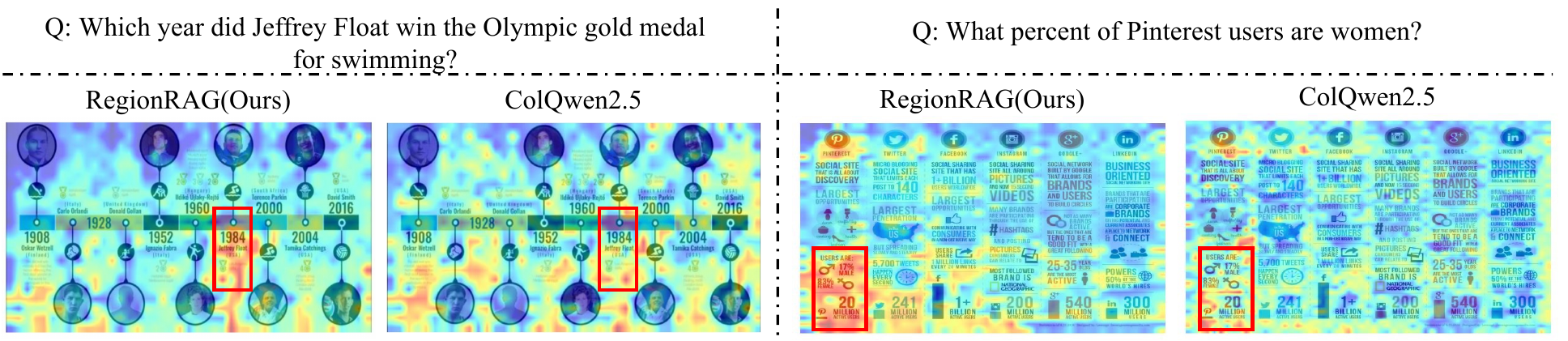} 
\caption{Visualization of similarity maps. The relevant regions are highlighted with red boxes for better reading. Our RegionRAG obtains better localization ability, as it exhibits higher and more concentrated similarity in relevant regions.}
\label{visal}
\end{figure*}

\begin{figure}[!t]
\centering
\includegraphics[width=0.95\columnwidth]{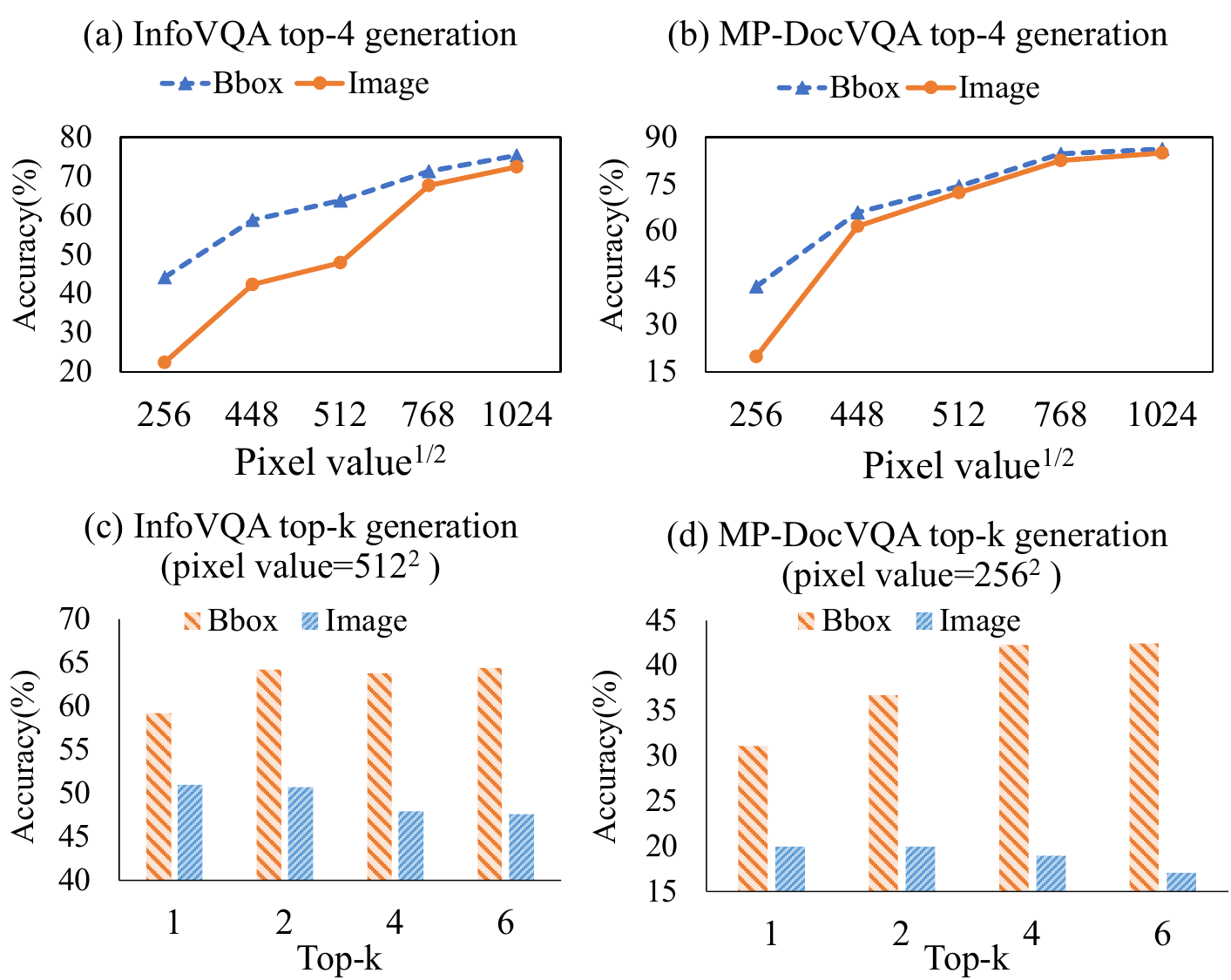}
\caption{Ablation study on generation performance. We compare using Bbox versus Image as input to the generator. (a) and (b) show accuracy versus input resolution for a fixed~$k=4$. (c) and (d) show accuracy as a function of top-$k$ candidates at a fixed resolution.}
\label{fig_gen}
\end{figure}

\subsection{Ablation}
\subsubsection{Main Ablation.}

\begin{table}[!t]
\small
\setlength\tabcolsep{1mm}
\centering
{%
\begin{tabular}{@{}l|clclcl@{}}
\toprule
\multicolumn{1}{l|}{\multirow{2}{*}{\textbf{Methods}}} & \multicolumn{2}{c}{\textbf{ArxivQA}} & \multicolumn{2}{c}{\textbf{DocVQA}} & \multicolumn{2}{c}{\textbf{SlideVQA}} \\
\multicolumn{1}{c|}{} & R@1 & N@5 & R@1 & N@5 & R@1 & N@5 \\ \midrule
RegionRAG (Ours) & \textbf{78.43} & \textbf{84.52} & \textbf{86.80} & \textbf{93.10} & \textbf{80.58} & \textbf{90.29} \\ \midrule
\hspace{3mm} w/o $\mathcal{L}_{local}$ & 77.08 & 83.41 & 84.94 & 91.70 & 78.24 & 88.90 \\
\hspace{3mm} w/o $\mathcal{L}_{global}$ & 77.57 & 81.68 & 85.45 & 81.14 & 78.24 & 86.85 \\ \midrule
\hspace{3mm} w/o Labeled data & 59.31 & 84.24 & 82.40 & 90.88 & 77.34 & 89.19 \\
\hspace{3mm} w/o Unlabeled data & 73.90 & 67.75 & 71.07 & 90.83 & 75.00 & 88.18 \\ \bottomrule
\end{tabular}%
}
\caption{Main ablation study (\%) of retrieval results for global contrastive learning ($\mathcal{L}_{global}$), regional contrastive learning ($\mathcal{L}_{local}$) with labeled and unlabeled data.}
\label{tab:ablation}
\end{table}
The core components of our method include the global contrastive loss ($\mathcal{L}_{global}$), the regional contrastive loss ($\mathcal{L}_{local}$), and a semi-supervised strategy that utilizes both labeled and unlabeled data. As shown in the Table~\ref{tab:ablation}, removing either~$\mathcal{L}_{global}$ or~$\mathcal{L}_{local}$ consistently leads to performance degradation across all datasets, indicating that both global document-level and fine-grained region-level contrastive learning are crucial for effective retrieval. Moreover, the results underscore the effectiveness of our semi-supervised data strategy. While discarding labeled data causes a substantial performance collapse, excluding unlabeled data also yields a notable decline (e.g., a 4.53\% drop in R@1 on ArxivQA). These findings confirm that our model not only relies on the supervised signal but also effectively leverages large-scale unlabeled data to enhance its generalization and robustness.

\subsubsection{Ablation Study on Generation Inputs.}
We conduct an ablation study to analyze the impact of input granularity on the generator. Specifically, we compare providing cropped bounding box (Bbox) regions against full image pages, focusing on two factors: input resolution (i.e., visual token count) and the number of top-$k$ retrieved candidates. As shown in Figure~\ref{fig_gen}(a) and (b), when inputs are constrained to a fixed resolution, the Bbox method consistently outperforms the Image method. The advantage is most pronounced at lower resolutions like~$256^2$ and~$448^2$, where Bbox accuracy exceeds Image by up to 21.73 points. This improvement arises because resizing a small Bbox crop enlarges its informative content compared to a down-sampled full image. Consequently, our approach achieves higher performance with fewer tokens (e.g., Bbox at~$256^2$ rivals Image at~$448^2$ on InfoVQA). At very high resolution ($1024^2$), the gap narrows as Qwen2.5-VL already localizes answers in high-fidelity images. Overall, our method strikes a better balance between performance and efficiency. Figure~\ref{fig_gen}(c) and (d) further analyze the effect of the top-$k$ setting at a fixed resolution. For all $k$, Bbox accuracy remains consistently higher than Image accuracy. Critically, as $k$ increases, Bbox performance improves while Image performance degrades. This contrast likely stems from attention dilution in the full-image setting, which introduces redundant and query-irrelevant information that distracts the LLM. In contrast, for the Bbox method, a larger $k$ presents an opportunity to provide more regions that are relevant to the query, leading to improved accuracy. These results confirm that our region-based approach effectively leverages multiple inputs, whereas the image-level approach is penalized by them.

\subsection{Qualitative Analysis}
To qualitatively investigate how RegionRAG achieves its strong performance, we visualize the similarity heatmaps in Figure~\ref{visal}. The figure clearly shows where the model gives the relevant regions to retrieve from a given query, comparing our method with ColQwen2.5, which processes the entire image. The similarity map of RegionRAG is significantly more concentrated on the query-relevant visual parts. In contrast, ColQwen2.5 focuses diffusely across the image, indicating a difficulty in distinguishing important information from background noise. For example, for the query about Jeffrey Float, our model pinpoints the exact `1984' entry on the timeline. Similarly, for the other example, RegionRAG precisely pinpoints the relevant charts and data points. This shows that our region-level retrieval mechanism effectively guides the model to the most salient information, which is key to its superior performance.

\section{Conclusion}
In this work, we presented RegionRAG, an innovative framework that enhances visual document RAG by retrieving fine-grained semantic regions instead of entire documents. By leveraging a dual-objective training strategy and a neighbor-based region grouping algorithm, RegionRAG effectively filters irrelevant context and provides the generator with precise visual evidence. Our method achieves state-of-the-art performance across six benchmarks, significantly improving both retrieval (+10.02\% in R@1) and question answering accuracy (+3.56\%) while cutting visual token costs by 28.58\%. In the future, we aim to build more efficient and scalable RAG systems, extending to more general scenarios.

\section*{Acknowledgments}
This work is supported by the National Nature Science Foundation of China (62425114, 62121002, U23B2028, 62272436). We thank the support of the GPU cluster built by MCC Lab of Information Science and Technology Institution, USTC, and USTC super-computing center for providing computational resources for this project.

\bibliography{main}


\input{appendix.tex}

\end{document}

%% file: appendix.tex
\appendix
\clearpage

\setcounter{page}{1}
\setcounter{table}{0}
\setcounter{figure}{0}
\setcounter{equation}{0}
\setcounter{footnote}{0}
\setcounter{algorithm}{0}
\renewcommand{\thetable}{S\arabic{table}}
\renewcommand{\thefigure}{S\arabic{figure}}
\renewcommand{\theequation}{S\arabic{equation}}
\renewcommand{\thealgorithm}{S\arabic{algorithm}}

\twocolumn[\begin{center}
    \LARGE
    \textbf{Technical Appendix}
    \vspace{1.0em}
\end{center}]

\subsubsection{Overview.}
In the Appendix, we introduce more methods details in Sec.~\ref{sec:methods}, more implementation details in Sec.~\ref{sec:implementation}, more dataset details in our training stage in Sec.~\ref{sec:new_data_stage}. Then we add more experiments in Sec.~\ref{sec:more_exp}, such as hyper-parameter studies, inference efficiency analysis.

\section{Methods Details}
\label{sec:methods}

\renewcommand{\algorithmicrequire}{\textbf{Input:}}
\renewcommand{\algorithmicensure}{\textbf{Output:}}
\begin{algorithm}[H]
\caption{Bounding Box Proposal from Saliency Map (Detailed)}
\begin{algorithmic}[1]
\Require Saliency map $S$, bbox threshold $\eta$, neighbor range $r$, image size $(H_{img}, W_{img})$, patch size $(H_p, W_p)$
\Ensure A set of bounding boxes $\mathcal{B}$

\State $M \leftarrow S \ge \eta$ \Comment{Binarize saliency map to get a mask}
\State $P_{salient} \leftarrow \text{get\_grid\_coordinates}(M)$ \Comment{Get 2D indices of salient patches}
\State $\mathcal{R} \leftarrow \emptyset$ \Comment{Initialize list of connected regions}
\State $visited \leftarrow \emptyset$ \Comment{Initialize set of visited patch coordinates}

\For{each patch coordinate $p \in P_{salient}$}
    \If{$p \notin visited$}
        \State $Q \leftarrow \text{collections.deque}()$ \Comment{Initialize a queue for BFS}
        \State $R_{current} \leftarrow \emptyset$ \Comment{Initialize the current region}
        
        \State $Q.\text{append}(p)$
        \State $visited.\text{add}(p)$
        
        \While{$Q$ is not empty}
            \State $p_{current} \leftarrow Q.\text{popleft}()$
            \State $R_{current}.\text{add}(p_{current})$
            
            \For{each neighbor $p_{neighbor}$ of $p_{current}$ within range $r$}
                \If{$p_{neighbor} \in P_{salient}$ and $p_{neighbor} \notin visited$}
                    \State $visited.\text{add}(p_{neighbor})$
                    \State $Q.\text{append}(p_{neighbor})$
                \EndIf
            \EndFor
        \EndWhile
        \State $\mathcal{R}.\text{add}(R_{current})$
    \EndIf
\EndFor

\State $\mathcal{B} \leftarrow \emptyset$ \Comment{Initialize the set of final bounding boxes}
\For{each region $R \in \mathcal{R}$}
    \State $(x_{min}, y_{min}) \leftarrow (\min_{p \in R} p.x, \min_{p \in R} p.y)$ \Comment{Find min grid coordinates}
    \State $(x_{max}, y_{max}) \leftarrow (\max_{p \in R} p.x, \max_{p \in R} p.y)$ \Comment{Find max grid coordinates}
    
    \State $x_1 \leftarrow x_{min} \times W_p$
    \State $y_1 \leftarrow y_{min} \times H_p$
    \State $x_2 \leftarrow \min((x_{max} + 1) \times W_p, W_{img})$
    \State $y_2 \leftarrow \min((y_{max} + 1) \times H_p, H_{img})$
    
    \State $B \leftarrow (x_1, y_1, x_2, y_2)$
    \State $\mathcal{B}.\text{add}(B)$
\EndFor

\State \textbf{return} $\mathcal{B}$
\end{algorithmic}
\label{BFS}
\end{algorithm}
To further clarify our method for extracting candidate regions from the saliency map (as described in Section 3.5 of the main text), this section provides a detailed algorithmic description. Algorithm~\ref{BFS} aims to effectively merge spatially adjacent patches with saliency scores above a threshold into meaningful semantic regions and to generate a minimum bounding box for each region. Specifically, the process begins by filtering for all patches with saliency scores exceeding a preset threshold $\eta$. It then employs a Breadth-First Search (BFS) to identify connected components, grouping adjacent patches that fall within a neighborhood range $r$. Once a connected region of patches is identified, the algorithm calculates its minimum bounding box by determining the extremal grid coordinates of all constituent patches and converting them into actual pixel coordinates on the image. This ultimately yields a set of precise candidate regions for the subsequent generator.

\section{Implementation Details}
\label{sec:implementation}

We initialized RegionRetriever with Qwen2.5-VL-3B~\cite{bai2025qwen2} for training, while the Generator is an off-the-shelf model without training. The model is trained using bfloat16 mixed-precision, and we leverage the flash-attention-2 implementation for efficiency. Instead of full fine-tuning, we employ Parameter-Efficient Fine-Tuning (PEFT) using the LoRA methodology. The LoRA configuration features a rank ($r$) of 32 and an $l_\alpha$ of 32, with a dropout rate of 0.1. LoRA adapters are applied to all linear projection layers. We used the AdamW optimizer with a learning rate of 2e-4 and a warmup phase of 100 steps. The temperature for the global contrastive loss ($\tau$) is set to 0.02, while the temperature of the local contrastive loss is 0.25. Given the global loss weighted by ~$\alpha=1$, the local loss component was weighted by a coefficient~$\beta=0.01$ in the final loss calculation.

\section{Dataset Details}
\label{sec:new_data_stage}
\begin{table*}[t]
\centering
{%
\begin{tabular}{llcl}
\toprule
\textbf{Domain} & \textbf{Source Dataset} & \textbf{Size} & \textbf{Dataset Description} \\ \midrule
\textbf{Unlabeled data} & VisRAG-in-domain & 122k & Diverse document images with text  \\ \midrule
\multirow{4}{*}{\centering \textbf{Labeled data}} & Visual-CoT TextVQA & 18k & Images with text \\
 & Visual-CoT TextCaps & 32k & Images with text \\
 & Visual-CoT DocVQA & 33k & Document images \\
 & Visual-CoT InfographicsVQA & 15k & Infographic \\
\bottomrule
\end{tabular}%
}
\caption{The overview of the document-focused subset of Visual-CoT. The dataset includes four source datasets.}
\label{tab:visualcot}
\end{table*}

\begin{figure*}[!t]
\centering
\includegraphics[width=0.84\textwidth]{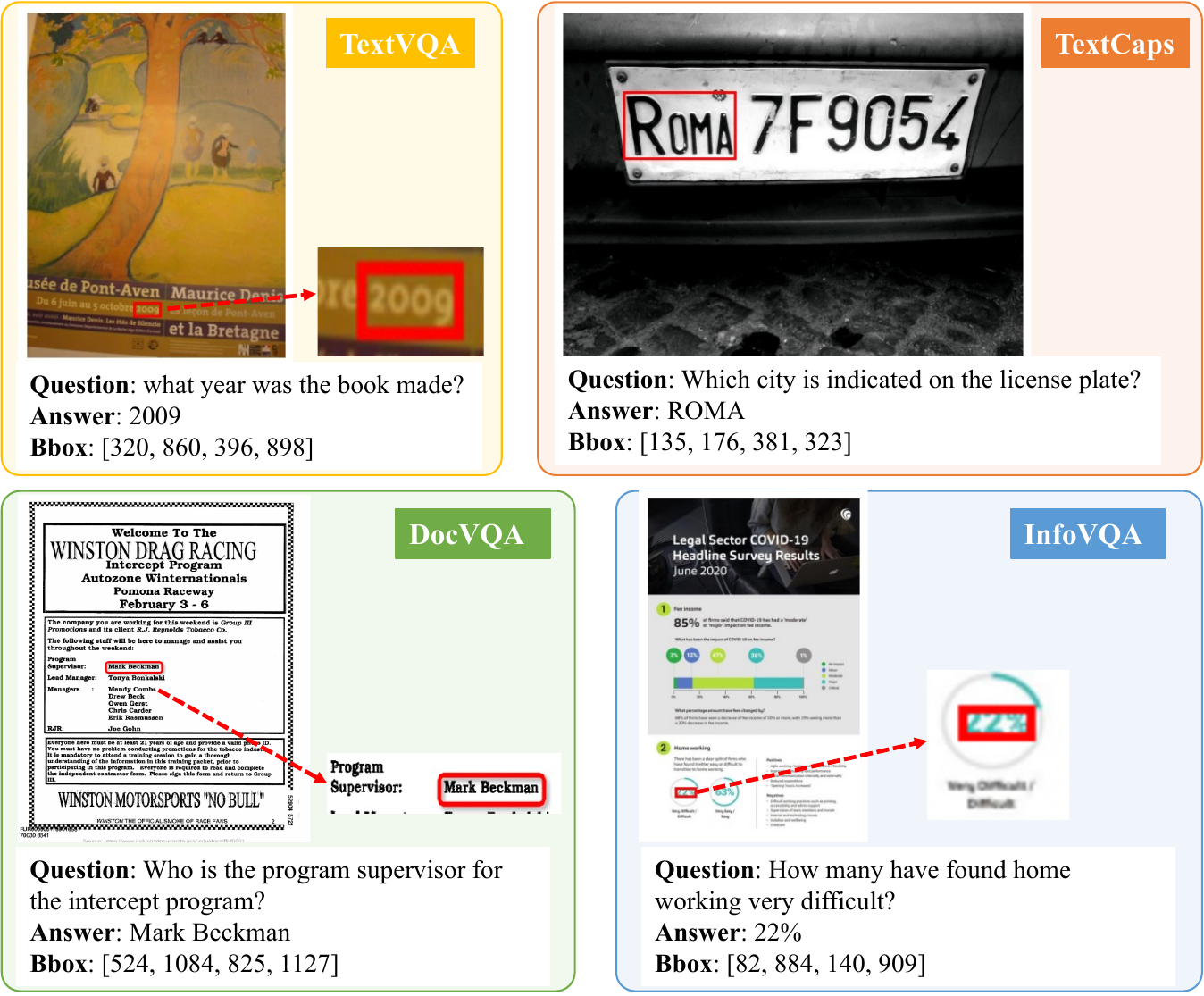} 
\caption{Examples of four sources covered in the Visual-CoT dataset, with corresponding question-answer annotations and bboxes. The red bounding boxes in the images highlight the critical image regions that provide necessary and related information for answering the questions.}
\label{visal}
\end{figure*}

Our model is trained on a hybrid dataset composed of both unlabeled and labeled data, with statistics detailed in Table~\ref{tab:visualcot}. For unlabeled data training, we utilize the VisRAG~\cite{yu2024visrag} in-domain dataset, which covering document types from scientific figures to industrial documents, is crucial for enhancing our model's generalization capability. For fine-grained supervision, the labeled data is sourced from Visual-CoT, which is composed of several benchmarks, including TextVQA~\cite{singh2019towards}, TextCaps~\cite{sidorov2020textcaps}, DocVQA~\cite{mathew2021docvqa}, and InfographicsVQA~\cite{mathew2022infographicvqa}.

Our ablation studies (as shown in Table 5) reveal the complex and distinct contributions of each data type, as the impact of removing them varies across datasets. For instance, on DocVQA, removing unlabeled data causes a more significant drop in R@1 (from 86.80\% to 71.07\%) than removing labeled data (to 82.40\%). Conversely, on ArxivQA, the effect is reversed and even more pronounced: removing labeled data leads to a dramatic drop in R@1 (from 78.43\% to 59.31\%), far greater than the impact of removing unlabeled data (to 73.90\%). This complex interaction suggests that the value of labeled data is not simply to boost a retrieval score, but to provide an invaluable and unique type of supervision signal. Specifically, tasks in datasets like DocVQA and TextVQA (e.g., ``What year was the book made?") require the model to directly analyze textual content from the image, rather than retrieving answers from an external knowledge base. Therefore, while such samples may not be ideal for training a conventional retrieval module, they are highly beneficial for guiding our model to accurately locate question-relevant information within the visual content, which is central to our RegionRAG framework.

\begin{table*}[t]
\centering
\begin{tabular}{@{}l|cccccccc@{}}
\toprule
\multirow{1}{*}{\textbf{Methods}} & \textbf{R@1} & \textbf{R@2} & \textbf{R@5} & \textbf{R@10} & \textbf{N@1} & \textbf{N@2} & \textbf{N@5} & \textbf{N@10}   \\ \midrule
 & \multicolumn{8}{l}{\small \textit{ArxivQA}} \\
VisRAG & 69.12 & 76.59 & 83.46 & 87.62 & 69.12 & 73.83 & 76.98 & 78.36 \\
ColQwen2.5 & 62.25 & 70.83 & 80.27 & 84.56 & 62.25 & 67.67 & 71.88 & 73.27 \\
RegionRAG(Ours) & \textbf{78.43} & \textbf{83.33} & \textbf{89.95} & \textbf{92.52} & \textbf{78.43} & \textbf{81.52} & \textbf{84.52} & \textbf{85.35}\\ \midrule
 & \multicolumn{8}{l}{\small \textit{DocVQA}} \\
VisRAG & 66.16 & 77.33 & 86.97 & 91.20 & 66.16 & 73.20 & 77.72 & 79.13 \\
ColQwen2.5 & 80.20 & 86.63 & 94.59 & 97.46 & 80.20 & 84.26 & 87.89 & 88.85 \\
RegionRAG(Ours) & \textbf{86.63} & \textbf{92.89} & \textbf{98.14} & \textbf{99.32} & \textbf{86.63} & \textbf{90.58} & \textbf{92.98} & \textbf{93.39} \\ \midrule
 & \multicolumn{8}{l}{\small \textit{InfoVQA}}  \\
VisRAG & 78.83 & 89.83 & 95.68 & 97.08 & 78.83 & 85.77 & 88.47 & 88.91 \\
ColQwen2.5 & 82.31 & 89.97 & 96.94 & 98.61 & 82.31 & 87.15 & 90.31 & 90.85 \\
RegionRAG(Ours) & \textbf{89.00} & \textbf{95.96} & \textbf{99.03} & \textbf{99.44} & \textbf{89.00} & \textbf{93.39} & \textbf{94.81} & \textbf{94.95} \\ \midrule
 & \multicolumn{8}{l}{\small \textit{PlotQA}}  \\
VisRAG & 49.13 & 61.88 & 80.76 & 68.38 & 49.13 & 57.17 & 65.54 & 68.38 \\
ColQwen2.5 & 9.27 & 14.83 & 28.62 & 47.51 & 9.27 & 12.78 & 18.96 & 25.05 \\
RegionRAG(Ours) & \textbf{54.69} & \textbf{68.13} & \textbf{83.08} & \textbf{92.35} & \textbf{54.69} & \textbf{63.17} & \textbf{69.84} & \textbf{72.88} \\ \midrule
 & \multicolumn{8}{l}{\small \textit{SlideVQA}}  \\
VisRAG & 76.35 & 87.77 & 95.05 & 97.12 & 79.01 & 87.09 & 89.90 & 90.19 \\
ColQwen2.5 & 76.08 & 87.95 & 94.60 & 96.76 & 76.08 & 83.57 & 86.74 & 87.44 \\
RegionRAG(Ours) & \textbf{80.04} & \textbf{91.37} & \textbf{97.30} & \textbf{98.20} & \textbf{80.04} & \textbf{87.18} & \textbf{90.01} & \textbf{90.30} \\
\bottomrule
\end{tabular}
\caption{Overall retrieval performance in Recall@$k$ and nDCG@$k$ for $k=1,2,5,10$.}
\label{retrie}
\end{table*}


\begin{figure}[!t]
\centering
\includegraphics[width=0.95\columnwidth]{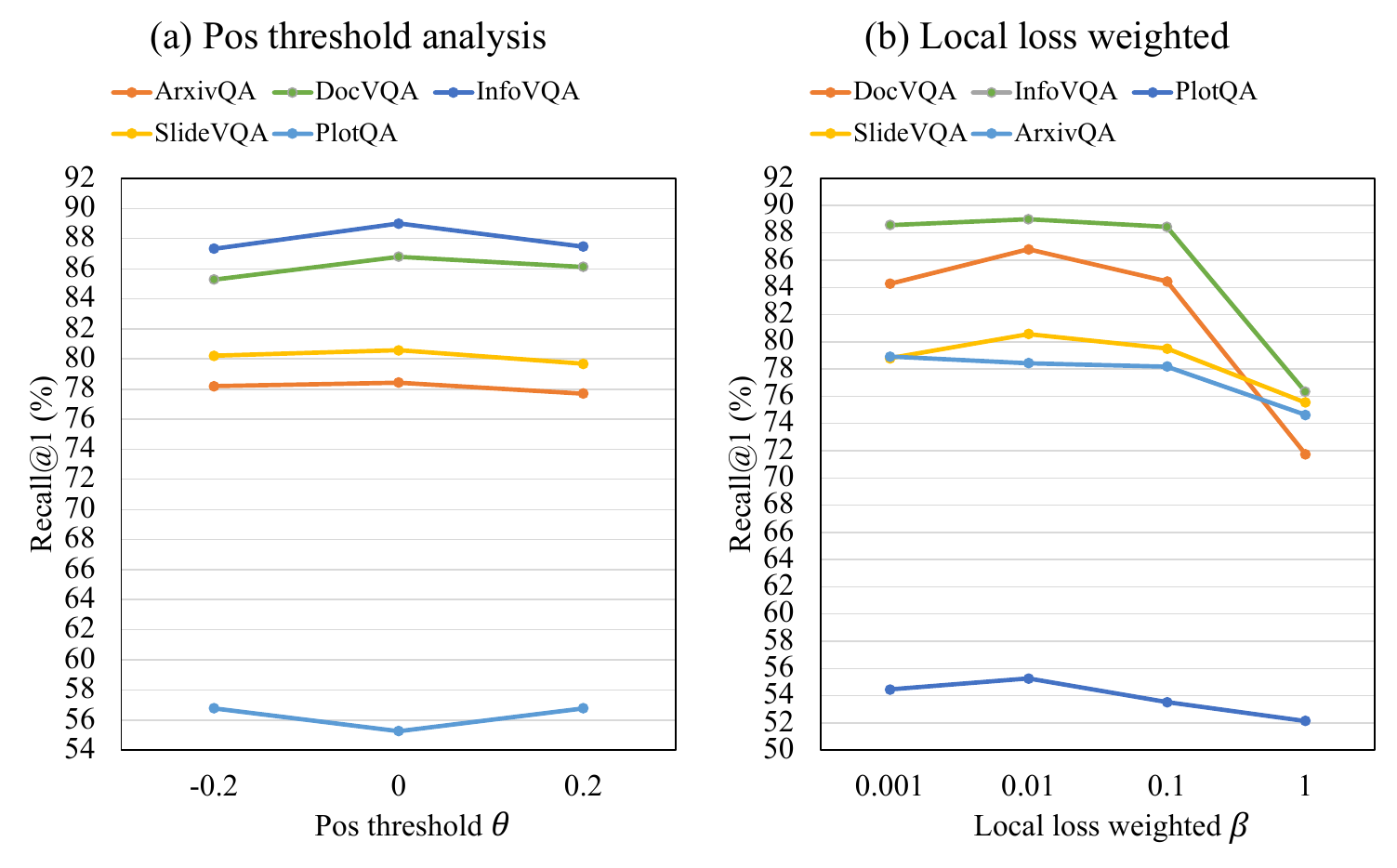} 
\caption{Analysis of the impact of key hyper-parameters on retrieval performance (Recall@1). (a) The effect of the positive pseudo-label threshold ($\theta$), used for training on unlabeled data, on Recall@1. (b) The effect of the local loss weight ($\beta$) on Recall@1. The results indicate that the model achieves optimal performance when $\theta=0$ and $\beta=0.01$.}
\label{1_hyper}
\end{figure}

\section{Additional Experimental Results}
\label{sec:more_exp}

\subsection{Retrieval Performance}


To more comprehensively evaluate the retrieval performance of our proposed RegionRAG model, we provide more detailed experimental results in Table~\ref{retrie}, expanding upon Table 1 and Table 2 in the main paper by comparing our method against the state-of-the-art baselines VisRAG~\cite{yu2024visrag} and ColQwen2.5~\cite{faysse2024colpali} across five standard visual document understanding benchmarks. The comparison covers a broader range of retrieval metrics, including Recall@$k$ (R@$k$) and nDCG@$k$ (N@$k$), for $k=\{1,2,5,10\}$. 


As shown in Table~\ref{retrie}, RegionRAG consistently outperforms both VisRAG and ColQwen2.5 across all metrics on all five benchmarks, which demonstrates the effectiveness and robustness of our proposed region-level retrieval paradigm. For instance, our model achieves an R@10 score of 92.35, whereas ColQwen2.5 scores only 47.51. This suggests that when information is highly localized and concentrated within a document (as in scientific plots), traditional document-level methods falter, while our region-aware approach can accurately pinpoint and leverage this information, showing an overwhelming performance advantage. 

Additionally, the consistent lead in the nDCG@$k$ metrics is particularly important. It indicates that RegionRAG not only succeeds at retrieving the correct document but, more crucially, is better at ranking it with higher confidence at the top of the list. For example, on the DocVQA dataset, our nDCG@10 score of 93.39 is substantially higher than ColQwen2.5's 88.85. This shows that the list of results retrieved by our method is of higher quality and more beneficial for the downstream generator.

\subsection{Hyper-parameter Analysis}


\subsubsection{Analysis of Positive Pseudo-label Threshold~$\theta$.} This threshold is critical for automatically partitioning pseudo-positive patches from the similarity map when training on unlabeled data, directly impacting the model's retrieval performance. Figure (a) illustrates the impact of this threshold on the model's retrieval performance (Recall@1) across four datasets. We tested three values: -0.2, 0, and 0.2. The results clearly show that Recall@1 most benchmarks peaks when the threshold is set to 0. This aligns with our intuition: a threshold that is too low (e.g., -0.2) might introduce excessive noise by incorrectly labeling irrelevant patches as positive. Conversely, a threshold that is too high (e.g., 0.2) could filter out too many ``hard" but relevant positive examples, weakening the supervisory signal. Therefore, setting the threshold to 0 achieves the best trade-off between the quality and quantity of pseudo-labels, maximizing retrieval performance.

\subsubsection{Analysis of Local Loss Weight~$\beta$.} The weight coefficient~$\beta$ balances the global document-level alignment loss and the local region-level alignment loss in our overall objective function. To investigate its effect, we show the model's retrieval performance (Recall@1) for different values of~$\beta$ in Figure (b). The results demonstrate that Recall@1 is optimal or near-optimal at~$\beta=0.01$ offers a robust and consistently high Recall@1 across datasets. While some datasets like InfoVQA and SlideVQA achieve a slightly higher peak at~$\beta=0.1$, that value is adjacent to a sharp performance decline at~$\beta=1$. When~$\beta$ is too small (e.g., 0.001), the model does not receive a sufficient local supervisory signal to learn fine-grained region localization. Conversely, when~$\beta$ is too large (e.g., 1), the model over-emphasizes local details at the potential expense of holistic document understanding, leading to a sharp decline in retrieval performance. We, therefore, select ~$\beta=0.01$, as it provides a well-calibrated regularization that enables the model to learn both macro and micro-level alignments synergistically for the best retrieval performance.

\subsubsection{Analysis of Neighbor Range.}
\begin{figure}[!t]
\centering
\includegraphics[width=0.95\columnwidth]{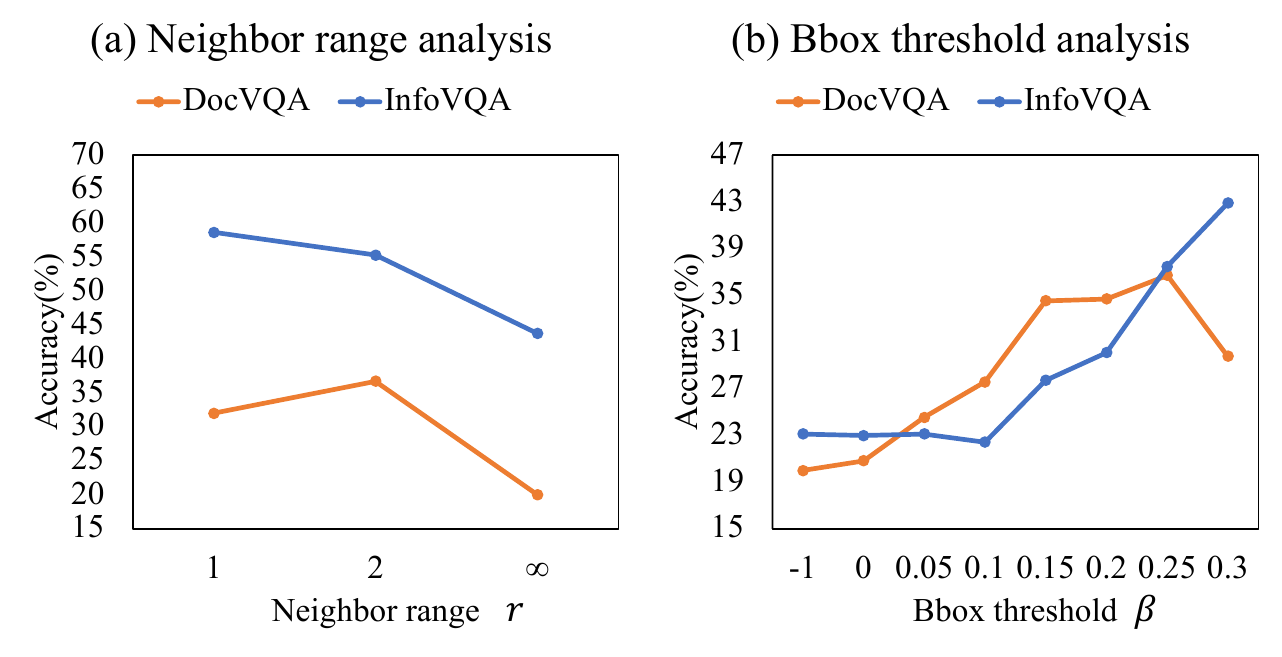} 
\caption{Hyper-parameter analysis. (a) The effect of neighbor range on accuracy. (b) The effect of bbox threshold on accuracy. A bbox threshold of -1 is equivalent to using the full image as input.}
\label{hyper1}
\end{figure}
Figure~\ref{hyper1}(a) illustrates the impact of the neighbor range hyper-parameter on generation performance. To ensure a fair comparison, we use a fixed resolution, a constant number of top-$k$ images, and a fixed bbox threshold for two distinct benchmarks. As defined in our methodology, a neighbor range approaching infinity ($\infty$) results in selecting the entire image. Since a range of 0 is trivial and the effects of a range of 3 or higher were empirically observed to be identical to infinity for most images, our analysis focuses on the critical values of 1, 2, and~$\infty$. For both benchmarks, performance is optimal when the neighbor range is set to 1, with accuracy declining as the range increases. This trend is attributable to the fact that a larger neighbor range leads to greater overlap between the selected bounding boxes. The resulting information redundancy appears to introduce noise and confuse the large language model, thereby degrading generation accuracy.

\subsubsection{Analysis of Bbox Threshold.} 
Figure~\ref{hyper1}(b) shows the influence of the bbox threshold on generation outcomes, analyzed on two benchmarks with a fixed neighbor range. The similarity map scores used for thresholding range from -1 to 1. Thus, setting the bbox threshold to -1 is equivalent to selecting the entire image. The results clearly show that for every bbox threshold value tested, the performance using extracted bounding boxes is superior to that of using the full image. Furthermore, the optimal bbox threshold is dataset-dependent. This variation arises because different benchmarks feature distinct document types and layouts, leading to varying levels of information density within the images. Consequently, the ideal threshold for isolating relevant content differs from one dataset to another.

\subsection{Long-context Experiment}

\begin{table}[t]
\centering
\begin{tabular}{lccc}
\toprule
\textbf{Model} & \textbf{R@1} & \textbf{R@2} & \textbf{R@10} \\
\midrule
ColQwen2.5     & 10.13 & 13.50 & 20.08 \\
RegionRAG      & \textbf{12.58} & \textbf{16.91} & \textbf{23.24} \\
\bottomrule
\end{tabular}
\caption{Comparison of retrieval results between RegionRAG and ColQwen2.5 on LongDocURL.}
\label{tab:long-context}
\end{table}
To further evaluate the robustness of RegionRAG in long-context scenarios, we conduct experiments on LongDocURL~\cite{deng2024longdocurl}, a benchmark involving multi-page (often $\ge$50) and lengthier document inputs. As shown in Table~\ref{tab:long-context}, RegionRAG consistently outperforms ColQwen2.5 across all retrieval metrics. These results demonstrate that RegionRAG effectively scales to long-document settings, where query-relevant content is more sparsely distributed and retrieval requires stronger discrimination across extensive visual contexts. The consistent gains confirm that the proposed region-level retrieval mechanism remains effective under longer input contexts, supporting its generalization beyond short and medium length document benchmarks.

\subsection{Inference Efficiency Analysis}
\begin{table}[t]
\centering
\begin{tabular}{lccc}
\toprule
\textbf{Method} & \textbf{Retrieval} & \textbf{Generation} & \textbf{Total} \\
\midrule
Image & 217.3 & 435.5 & 652.8 \\
Bbox  & 309.6 & 291.1 & 600.7 \\
\bottomrule
\end{tabular}
\caption{Comparison of retrieval, generation, and total inference time (in seconds) between image-level and region-level methods on InfoVQA.}
\label{tab:efficiency}
\end{table}
To further assess the computational efficiency of RegionRAG, we measure the inference latency on InfoVQA, comparing image-level (Image) and region-level (Bbox) pipelines. As shown in Table~\ref{tab:efficiency}, the Image approach incurs slightly higher retrieval time (309.6 s vs. 217.3 s) due to additional region selection and merging operations. However, this overhead is offset by a substantial reduction in generation time (291.1 s vs. 435.5 s), as the model processes fewer visual tokens and focuses on compact, query-relevant regions. Overall, the total inference time of RegionRAG is reduced from 652.8 s to 600.7 s, achieving both higher efficiency and better generation quality. These results indicate that region-level retrieval introduces only minor pre-processing costs while providing significant computational savings during generation, leading to a more efficient inference pipeline.

\subsection{Evaluation with GPT-4o as Generator}

\begin{table}[!t]
\setlength\tabcolsep{1.5mm}
\centering
\begin{tabular}{@{}clccccc@{}}
\toprule
\textbf{\small Pixel} & \multicolumn{1}{c}{\textbf{\small Methods}} & \textbf{\small Arxiv} & \textbf{\small Doc} & \textbf{\small Info} & \textbf{\small Plot} & \textbf{\small Slide} \\ \midrule
\multirow{5}{*}{\rotatebox{90}{256$^\text{2}$}} & top1 image & 60.17 & 34.01 & 38.85 & 17.17 & 46.76 \\
 & top1 bbox & \textbf{61.52} & \textbf{36.37} & \textbf{49.44} & \textbf{17.92} & \textbf{46.94} \\ \cmidrule(l){2-7} 
 & top4 image & 59.31 & 31.30 & 38.86 & 17.03 & 48.56 \\
 & top4 bbox & \textbf{59.44} & \textbf{45.69} & \textbf{54.17} & \textbf{17.69} & \textbf{49.11} \\ \cmidrule(l){2-7} 
 & oracle & 61.52 & 36.55 & 42.34 & 20.05 & 51.44 \\ \midrule
\multirow{5}{*}{\rotatebox{90}{512$^\text{2}$}} & top1 image & \textbf{59.31} & 65.31 & 52.08 & 17.38 & 53.59 \\
 & top1 bbox & 59.21 & \textbf{73.44} & \textbf{56.55} & \textbf{18.08} & \textbf{54.16} \\ \cmidrule(l){2-7}
 & top4 image & 58.82 & 65.99 & 48.05 & 16.69 & 55.58 \\
 & top4 bbox & \textbf{60.42} & \textbf{76.64} & \textbf{49.86} & \textbf{17.57} & \textbf{58.09} \\ \cmidrule(l){2-7}
 & oracle & 63.24 & 70.89 & 55.71 & 24.33 & 56.12 \\ \bottomrule
\end{tabular}%
\caption{Performance comparison under fixed and dynamic resolutions across five VQA benchmarks (GPT-4o as generator). ``topk image/bbox" denotes processing the retrieved full image or bounding box, ``oracle" uses ground-truth images.}
\label{tab:gpt4o}
\end{table}
To verify the generalizability of our approach beyond Qwen2.5-VL, we further conduct experiments by replacing the generator with GPT-4o while keeping the same retrieval pipeline. As shown in Table~\ref{tab:gpt4o}, our region-level input strategy consistently outperforms the image-level counterpart across all benchmarks and resolutions. The advantage is especially evident at lower resolutions (e.g., $256^2$), where the Bbox-based input achieves up to 10.6 points higher accuracy on InfoVQA. This demonstrates that our region-aware retrieval mechanism provides clear and focused visual grounding that benefits even a powerful generator like GPT-4o. Moreover, the performance gain remains stable at higher resolutions ($512^2$), indicating that the effectiveness of RegionRAG is model-agnostic and transfers well to stronger large multi-modal models. These results confirm that the improvements brought by our method stem from better input organization rather than being tied to a specific generator architecture.

\subsection{Fine-tuning Baselines on Our Dataset}
\begin{table}[!t]
\begin{tabular}{lllllll}
\toprule
\textbf{Model} & \textbf{data} & \textbf{Arxiv} & \textbf{Doc} & \textbf{Info} & \textbf{Plot} & \textbf{Slide} \\ \midrule
\multirow{2}{*}{visrag} & 362k & 87.62 & 91.20 & 97.08 & 68.38 & 97.12  \\
 & 220k & 81.31 & 92.12 & 95.33 & 61.87 & 96.86  \\ \midrule
Ours & 220k & \textbf{92.52} & \textbf{99.32} & \textbf{99.44} & \textbf{92.35} & \textbf{98.20} \\ \bottomrule
\end{tabular}
\caption{Retrieval performance (Recall@10) comparison between RegionRAG and fine-tuned baseline (VisRAG) on our 220k training set.}
\label{tab:visrag_finetune}
\end{table}
To ensure a fair comparison, we further fine-tune VisRAG and ColQwen2.5 on our 220k training set and compare their retrieval performance. ColQwen2.5 shares the same backbone as our base model (Qwen2.5-VL-3B), and its fine-tuning configuration corresponds to the ``w/o local loss" setting in our ablation study (Table~\ref{tab:ablation}). As shown in Table~\ref{tab:visrag_finetune}, VisRAG exhibits a noticeable performance drop when trained on our smaller dataset (220k vs. its original 362k samples). In contrast, RegionRAG consistently outperforms both fine-tuned VisRAG and ColQwen2.5 across all benchmarks, achieving substantial improvements particularly on the PlotQA and ArxivQA datasets. These results demonstrate that the superior performance of RegionRAG stems primarily from our proposed regional contrastive learning and retrieval strategy, rather than from differences in data scale or fine-tuning setup.

\subsection{More Qualitative Analysis}
\begin{figure*}[]
\centering
\includegraphics[width=0.95\textwidth]{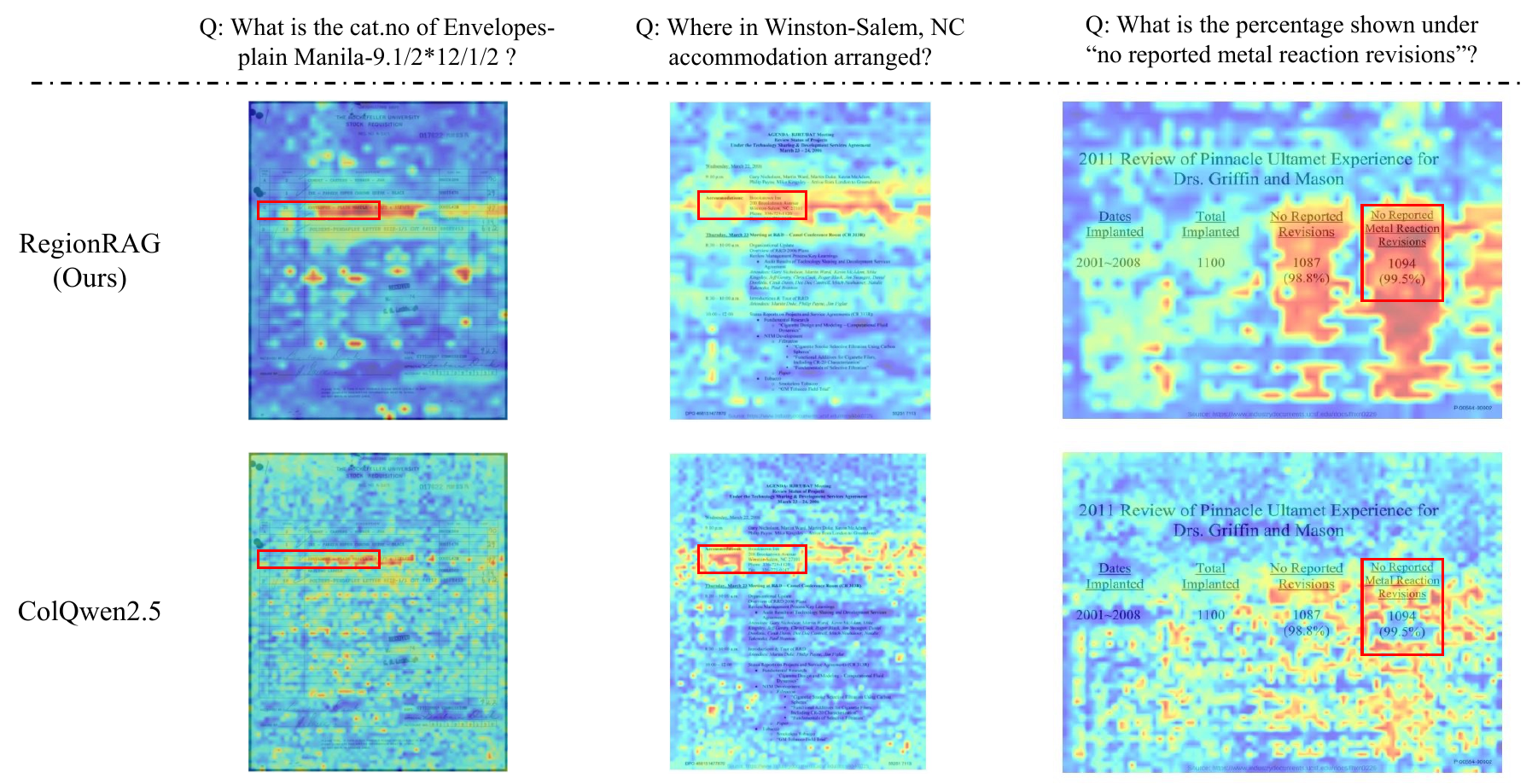} 
\caption{Qualitative comparison of similarity heatmaps on the DocVQA dataset. For various questions, our RegionRAG model generates significantly more focused and intense activation heatmaps on the ground-truth regions (indicated by red boxes) compared to the more diffuse heatmaps from the ColQwen2.5 baseline.}
\label{docvqa}
\end{figure*}

\begin{figure*}[]
\centering
\includegraphics[width=0.9\textwidth]{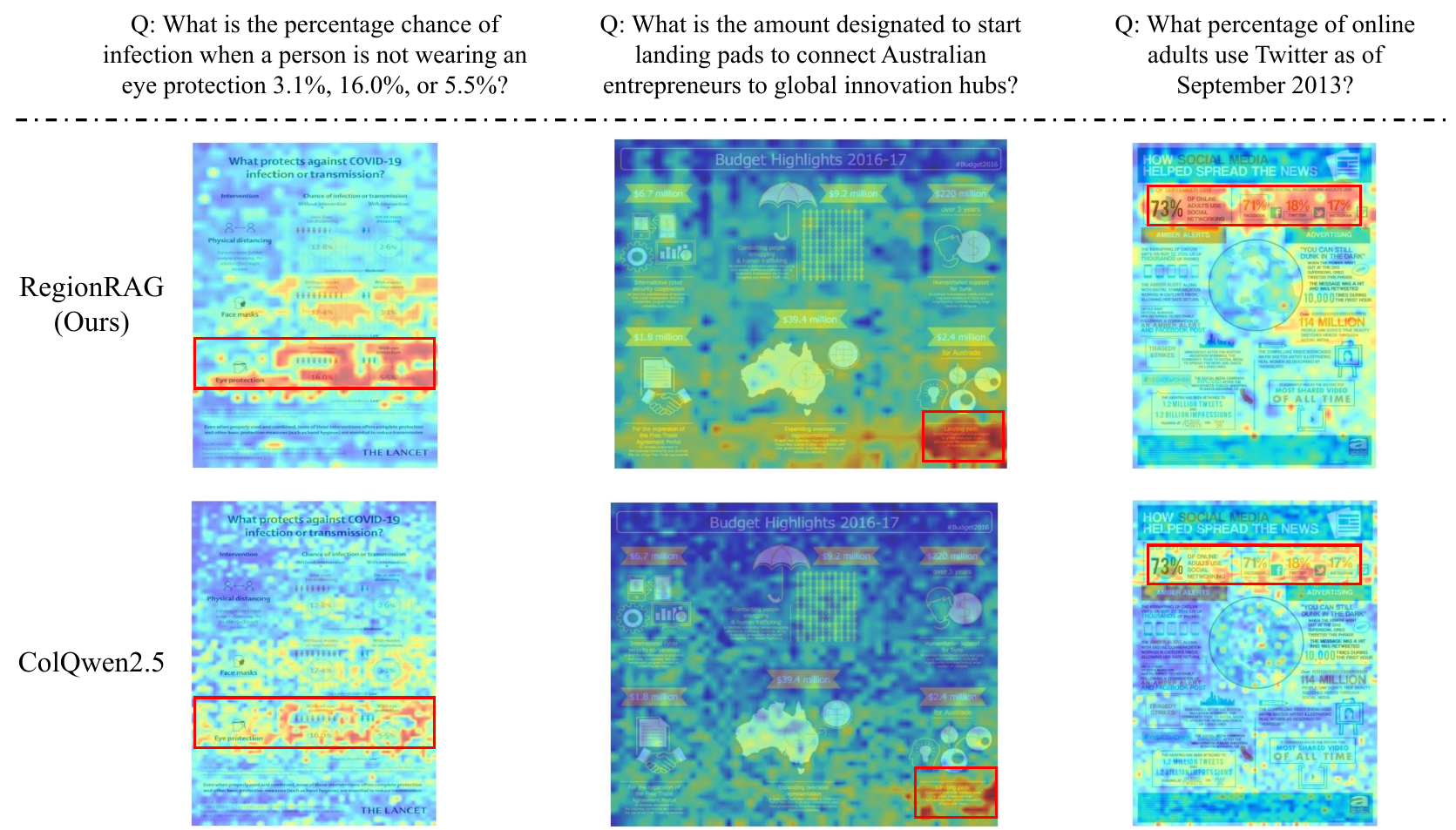} 
\caption{Qualitative comparison of similarity heatmaps on the InfoVQA dataset. Similar to its performance on DocVQA, RegionRAG demonstrates superior localization ability on infographics, generating sharp heatmaps that are precisely focused on small, answer-bearing regions (indicated by red boxes), while the baseline's attention is more scattered.}
\label{infovqa}
\end{figure*}

To intuitively demonstrate the superior localization ability of our RegionRAG model, we provide a qualitative comparison of similarity heatmaps against the ColQwen2.5 baseline on the DocVQA and InfoVQA datasets in Figure~\ref{docvqa} and Figure~\ref{infovqa}. These visualizations clearly show that for a given question, RegionRAG generates highly focused and intense ``hotspots" that precisely cover the text or chart area containing the answer. For instance, in a table-based question from DocVQA (Figure~\ref{docvqa}, fourth from left), RegionRAG accurately focuses its attention on the specific cell corresponding to ``no reported metal reaction revisions", whereas the attention from ColQwen2.5 is comparatively diffuse and fails to precisely lock onto the information. This precise localization capability is a direct result of our proposed region-level contrastive learning objective, which enables the model to effectively filter irrelevant information and focus on the most critical evidence, thereby significantly boosting performance on complex visual document understanding tasks.

\subsection{Generalization Across Model Architectures}
\begin{table}[!t]
\centering
\begin{tabular}{lll}
\toprule
\textbf{Method} & \textbf{Bbox} & \textbf{Image} \\ \midrule
\textbf{Accuracy (\%)} & 33.1 & 20.7 \\ \bottomrule
\end{tabular}
\caption{Generation performance on InfoVQA using PaliGemma-3B as the retriever. The Bbox input still outperforms the Image input under the top-1 QA setting.}
\label{tab:paligemma_gen}
\end{table}

\begin{table*}[!t]
\centering
\begin{tabular}{llllllll}
\toprule
\textbf{Model} & \textbf{Retriever} & \textbf{ArxivQA} & \textbf{DocVQA} & \textbf{InfoVQA} & \textbf{PlotQA} & \textbf{SlideVQA} & \textbf{Avg.} \\ \midrule
ColPali & PaliGemma-3B & 79.41 & 90.18 & 88.86 & 31.41 & 92.81 & 76.53 \\ \midrule
\multirow{2}{*}{RegionRAG} & PaliGemma-3B & 82.72 & 95.26 & 95.54 & 85.05 & 97.12 & 91.14 \\
 & Qwen2.5-VL-3B & 92.52 & 99.32 & 99.44 & 92.35 & 98.20 & 96.37 \\ \bottomrule
\end{tabular}
\caption{Retrieval performance (Recall@10) comparison using different base models. RegionRAG maintains superior results over ColPali when both adopt the same PaliGemma-3B backbone.}
\label{tab:paligemma}
\end{table*}
\subsubsection{PaliGemma.}
PaliGemma~\cite{beyer2024paligemma} is a versatile 3B vision-language model built upon the Gemma language backbone and the Pali visual encoder. Despite its compact size, it demonstrates strong transferability across diverse multimodal tasks such as VQA and document understanding. In our experiments, we adopt PaliGemma-3B as the backbone to assess the generalization of RegionRAG across different model architectures (Table~\ref{tab:paligemma} and Table~\ref{tab:paligemma_gen}).

\subsubsection{Retrieval Performance.}
To evaluate the generalization of RegionRAG across different architectures, we replace the Qwen2.5-VL-3B backbone used in the main experiments with PaliGemma-3B, the same base model adopted by ColPali. As shown in Table~\ref{tab:paligemma}, RegionRAG still achieves significantly higher retrieval performance across all benchmarks, with an average Recall@10 improvement of 14.61 percentage points over ColPali. In particular, the gap is most pronounced on PlotQA, where RegionRAG surpasses ColPali by over 50 points (85.05 vs. 31.41). These results confirm that the advantages of our method do not stem from the backbone architecture, but rather from its region-level retrieval design and fine-grained contrastive learning strategy. This demonstrates that RegionRAG generalizes effectively across model architectures and scales, reinforcing the robustness and versatility of our framework.

\subsubsection{Generation Performance.}
To further examine whether the advantage of RegionRAG extends beyond retrieval, we evaluate its generation performance on InfoVQA using PaliGemma-3B as the backbone. As shown in Table~\ref{tab:paligemma_gen}, the model using retrieved bounding boxes achieves markedly higher accuracy (33.1\% vs. 20.7\%) compared to the full-image input under the top-1 QA setting. This result mirrors our earlier findings with Qwen2.5-VL, confirming that fine-grained, region-level inputs lead to more focused and efficient reasoning. Even with a different architecture, RegionRAG effectively preserves its performance advantage, demonstrating strong generalization of our region retrieval design across both model backbones and task stages.